\documentclass[letterpaper, 10 pt, conference]{IEEEtran}  

\IEEEoverridecommandlockouts                              

\usepackage{amsmath} 
\usepackage{amssymb}  
\usepackage{siunitx}
\usepackage{multirow}
\usepackage{xspace}
\usepackage{amsthm}

\usepackage{tikz}
\usepackage{pgfplots}
\usepackage{tikzscale}
\usepackage{marvosym}
\usepackage{balance}
\usepackage{lipsum}
\usepackage[caption=false]{subfig}
\usepackage{hyperref}

\makeatletter

\pgfplotsset{compat=1.18}
\usepackage{pgfplotstable}
\usepackage{pgfplots}
\usetikzlibrary{external}
\usetikzlibrary{matrix, decorations.pathreplacing, calc, positioning, fit, fadings, shapes.arrows, arrows.meta, shapes.geometric, backgrounds}
\usetikzlibrary{shadings}
\usetikzlibrary{fadings,decorations.pathreplacing}
\usetikzlibrary{shapes.arrows}
\usetikzlibrary{math}
\definecolor{custom-blue}{HTML}{1a80bb}
\definecolor{custom-red}{HTML}{a00000}
\definecolor{custom-yellow}{HTML}{f2c45f}
\definecolor{muted-gray}{HTML}{808080}
\definecolor{muted-gold}{HTML}{f0c571}
\definecolor{muted-teal}{HTML}{59a89c}
\definecolor{muted-blue}{HTML}{0b81a2}
\definecolor{muted-red}{HTML}{e25759}
\definecolor{muted-darkred}{HTML}{9d2c00}
\definecolor{muted-purple}{HTML}{7e4794}
\definecolor{muted-green}{HTML}{36b700}
\definecolor{muted-darkblue}{HTML}{082a54}
\definecolor{grid-green}{HTML}{67a05f}
\definecolor{grid-blue}{HTML}{0000ff}
\definecolor{mec-red}{HTML}{dc0000}

\definecolor{dark-red}{HTML}{c31e23}
\definecolor{light-red}{HTML}{ff5a5e}
\definecolor{med-brown}{HTML}{ea801c}
\definecolor{light-brown}{HTML}{f0b077}

\def\coloralpha{90}

\tikzstyle{arrow} = [thick,->,>=stealth]
\tikzstyle{doublearrow} = [thick,<->,>=stealth]
\tikzstyle{box} = [draw, rectangle, minimum width=1.9cm, minimum height=0.9cm, font=\footnotesize]

\def \minwidth {1.2cm}
\tikzstyle{rounded box} = [draw, rectangle, rounded corners, minimum height=.7cm, minimum width=\minwidth, font=\scriptsize]

\newif\iftikz@shading@path

\tikzset{
    fading xsep/.store in=\pgfpathfadingxsep,
    fading ysep/.store in=\pgfpathfadingysep,
    fading sep/.style={fading xsep=#1, fading ysep=#1},
    fading sep=0cm,
    shading path/.code={%
        \iftikz@shading@path%
        \else%
            \tikz@shading@pathtrue%
            \tikz@addmode{%
                \pgfgetpath\pgf@currentfadingpath%
                \pgfextract@process\pgf@fadingpath@southwest{\pgfpointadd{\pgfqpoint{\pgf@pathminx}{\pgf@pathminy}}%
                    {\pgfpoint{-\pgfpathfadingxsep}{-\pgfpathfadingysep}}}
                \pgfextract@process\pgf@fadingpath@northeast{\pgfpointadd{\pgfqpoint{\pgf@pathmaxx}{\pgf@pathmaxy}}%
                    {\pgfpoint{\pgfpathfadingxsep}{\pgfpathfadingysep}}}%
                \pgfsetpath\pgfutil@empty%
                \pgfinterruptpath%
                \pgfinterruptpicture%
                    \begin{tikzfadingfrompicture}[name=.]
                        \path [shade=none,fill=none, #1] \pgfextra{%
                            \pgfsetpath\pgf@currentfadingpath%
                            \pgf@fadingpath@southwest
                            \expandafter\pgf@protocolsizes{\the\pgf@x}{\the\pgf@y}%
                            \pgf@fadingpath@northeast%
                            \expandafter\pgf@protocolsizes{\the\pgf@x}{\the\pgf@y}%
                        };
                        \xdef\pgf@fadingboundingbox@southwest{\noexpand\pgfqpoint{\the\pgf@picminx}{\the\pgf@picminy}}%
                        \xdef\pgf@fadingboundingbox@northeast{\noexpand\pgfqpoint{\the\pgf@picmaxx}{\the\pgf@picmaxy}}%
                    \end{tikzfadingfrompicture}%
                \endpgfinterruptpicture%
                \endpgfinterruptpath%
                \pgfpathrectanglecorners{\pgf@fadingboundingbox@southwest}{\pgf@fadingboundingbox@northeast}%
                \def\tikz@path@fading{.}%
                \tikz@mode@fade@pathtrue%
                \tikz@fade@adjustfalse
                \pgfpointscale{0.5}{\pgfpointadd{\pgf@fadingboundingbox@southwest}{\pgf@fadingboundingbox@northeast}}%
                \edef\tikz@fade@transform{shift={(\the\pgf@x,\the\pgf@y)}}%
            }%
        \fi%
    }
}

\newcounter{groupcount}
\pgfplotsset{
    draw group line/.style n args={5}{
        after end axis/.append code={
            \setcounter{groupcount}{0}
            \pgfplotstableforeachcolumnelement{#1}\of\datatable\as\cell{%
                \def\temp{#2}
                \ifx\temp\cell
                    \ifnum\thegroupcount=0
                        \stepcounter{groupcount}
                        \pgfplotstablegetelem{\pgfplotstablerow}{pos}\of\datatable
                        \coordinate [yshift=#4] (startgroup) at (axis cs:\pgfplotsretval,0);
                    \else
                        \pgfplotstablegetelem{\pgfplotstablerow}{pos}\of\datatable
                        \coordinate [yshift=#4] (endgroup) at (axis cs:\pgfplotsretval,0);
                    \fi
                \else
                    \ifnum\thegroupcount=1
                        \setcounter{groupcount}{0}
                        \draw [
                            shorten >=-#5,
                            shorten <=-#5
                        ] (startgroup) -- node [anchor=base, yshift=0.5ex] {\scriptsize{#3}} (endgroup);
                    \fi
                \fi
            }
            \ifnum\thegroupcount=1
                        \setcounter{groupcount}{0}
                        \draw [
                            shorten >=-#5,
                            shorten <=-#5
                        ] (startgroup) -- node [anchor=base, yshift=0.5ex] {\scriptsize{#3}} (endgroup);
            \fi
        }
    }
}

\def\patchSize{1}
\def\patchSizeCm{0.15*\patchSize cm}
\def\gridMargin{0.15}
\newcommand{\drawPatch}[4][\patchSizeCm]{%

    \begin{scope}[%
            shift={#3},%
        ]
        \tikzmath{
            integer \startIdx;
            \startIdx= -#2/2;
            integer \endIdx;
            \endIdx = #2/2 - 1;
            \cellsize = (\patchSize - \gridMargin) / #2;
            function isCorner(\a,\b) {
                if \a==\startIdx then {
                    if \b==\startIdx then{ return 1;}
                    else {
                        if \b==\endIdx then{return 1;} 
                        else{ return 0;};
                    };
                } else{ 
                    if \a==\endIdx then {
                        if \b==\startIdx then{return 1;}
                        else {
                            if \b==\endIdx then{return 1;} 
                            else{return 0;};
                        };
                    } else {return 0;};
                };
            };
            {
                \draw[fill=white,rounded corners=#1, #4] (\startIdx*\cellsize, \startIdx*\cellsize) rectangle ++(#2*\cellsize, #2*\cellsize);
            };
            if #2==2 then{%
                {
                    \draw[#4] (-\cellsize,0) -- (\cellsize,0);
                    \draw[#4] (0,-\cellsize) -- (0,\cellsize);
                };
            } else{%
                integer \x;
                integer \y;
                for \x in {\startIdx ,..., \endIdx}{
                    for \y in {\startIdx ,...,\endIdx}{
                        if isCorner(\x,\y) then {} else {
                            {
                                \draw[fill=white, #4] (\x*\cellsize, \y*\cellsize) rectangle ++(\cellsize, \cellsize);
                            };
                        };
                    };
                };
            };
        }
    \end{scope}
}

\makeatletter
\newcommand\footnoteref[1]{\protected@xdef\@thefnmark{\ref{#1}}\@footnotemark}
\makeatother

\newcommand{\ros}{ROS~2\xspace}
\newcommand{\ete}{end-to-end\xspace}

\newcommand\HUGE{\@setfontsize\Huge{40}{50}}
\makeatother
\title{\LARGE \bf
Analysis of Efficient Transmission Methods of Grid Maps for Intelligent Vehicles 
}

\ifdefined\DOUBLEBLIND
  \author{Empty for double-blind review.}
\else
\author{Robin Dehler$^\dagger$, Dominik Authaler$^\dagger$, Aryan Thakur, Thomas Wodtko, and Michael Buchholz
\thanks{This work has been financially supported by the State Ministry of Economic Affairs Baden-Württemberg (project U-Shift II, AZ 3-433.62-DLR/60)} 
\thanks{All authors are with the Institute of Measurement, Control and Microtechnology, Ulm University, Albert-Einstein-Allee 41, 89081 Ulm, Germany {\tt\footnotesize \{firstname\}.\{lastname\}@uni-ulm.de}}%
\thanks{$\dagger$ R. Dehler and D. Authaler are both first authors with equal contribution.}	
}
\fi

\newcommand\copyrighttext{\footnotesize \textcopyright~2026 IEEE. Personal use of this material is permitted. Permission from IEEE must be obtained for all other uses, in any current or future media, including reprinting/republishing this material for advertising or promotional purposes, creating new collective works, for resale or redistribution to servers or lists, or reuse of any copyrighted component of this work in other works.%
}
\newcommand\copyrightnotice[1]{%
    \begin{tikzpicture}[remember picture,overlay]%
     \node[%
        anchor=south, %
        yshift=#1pt%
    ] at (current page.south)%
     {\fbox{\parbox{\dimexpr\textwidth-\fboxsep-\fboxrule\relax}{\copyrighttext}}};%
     \end{tikzpicture}%
}

\begin{document}

\maketitle
\thispagestyle{empty}
\pagestyle{empty}

%
%
%
%
%


\begin{abstract}
Grid mapping is a fundamental approach to modeling the environment of intelligent vehicles or robots.
Compared with object-based environment modeling, grid maps offer the distinct advantage of representing the environment without requiring any assumptions about objects, such as type or shape.
For grid-map-based approaches, the environment is divided into cells, each containing information about its respective area, such as occupancy.
This representation of the entire environment is crucial for achieving higher levels of autonomy.
However, it has the drawback that modeling the scene at the cell level results in inherently large data sizes.
Patched grid maps tackle this issue to a certain extent by adapting cell sizes in specific areas.
Nevertheless, the data sizes of patched grid maps are still too large for novel distributed processing setups or vehicle-to-everything (V2X) applications.
Our work builds on a patch-based grid-map approach and investigates the size problem from a communication perspective.
To address this, we propose a patch-based communication pipeline that leverages existing compression algorithms to transmit grid-map data efficiently.
We provide a comprehensive analysis of this pipeline for both intra-vehicle and V2X-based communication.
The analysis is verified for these use cases with two real-world experiment setups.
Finally, we summarize recommended guidelines for the efficient transmission of grid-map data in intelligent transportation systems. 
\end{abstract}
\copyrightnotice{15}
\section{Introduction}
Safety is considered to be of utmost importance for autonomous driving.
A reliable perception of the surrounding environment is key to safely executing driving tasks.
In recent years, intelligent vehicles have been equipped with an increasing number of sensors to meet robustness demands.
As a result, the amount of data that needs to be processed is growing significantly, and more computational power is required.

Two major strategies have evolved to provide sufficient computational power or to improve energy consumption.
First, multiple processing units are used and connected within intelligent vehicles, e.g., in~\cite{ushift2022, UNICAR}, sensor processing is handled by separate processing units.
Each unit can be designed to be task-specific and, therefore, energy efficient.
\ifdefined\DOUBLEBLIND
Second, infrastructural assistance is leveraged to provide preprocessed information by multiple traffic participants~\cite{vanKempen23}, and, more recently, function offloading methods have been proposed to dispense specific tasks to external processing units~\cite{saeik2021}.
\else
Second, infrastructural assistance is leveraged to provide preprocessed information by multiple traffic participants~\cite{vanKempen23, buchholz22}, and, more recently, function offloading methods have been proposed to dispense specific tasks to external processing units~\cite{saeik2021, dehler25}.
\fi
Regardless of the distribution of computing power, commonly used approaches for sensor fusion include object-based tracking~\cite{vo2013glmb, reuter2014lmb} and grid-map-based~\cite{thrun2005probabilistic} fusion methods.
While both supplement each other and provide valuable information about the environment, a significant difference becomes apparent in distributed processing, namely, data size.

\ifdefined\DOUBLEBLIND
Object-based methods are well established in infrastructure-aided perception, and respective standards exist~\cite{etsi_tr_103_562_intelligent_2019, etsi_en_302_637-2_intelligent_2014}.
\else
Object-based methods are well established in infrastructure-aided perception~\cite{buchholz22}, and respective standards exist~\cite{etsi_tr_103_562_intelligent_2019, etsi_en_302_637-2_intelligent_2014}.
\fi
For grid-map data, however, the sizes to be transmitted must be reduced to achieve low latencies for intra-vehicle transmission and to enable grid-map-based fusion approaches in combination with infrastructure support in the first place.
\begin{figure}[t]
    \centering
    \input{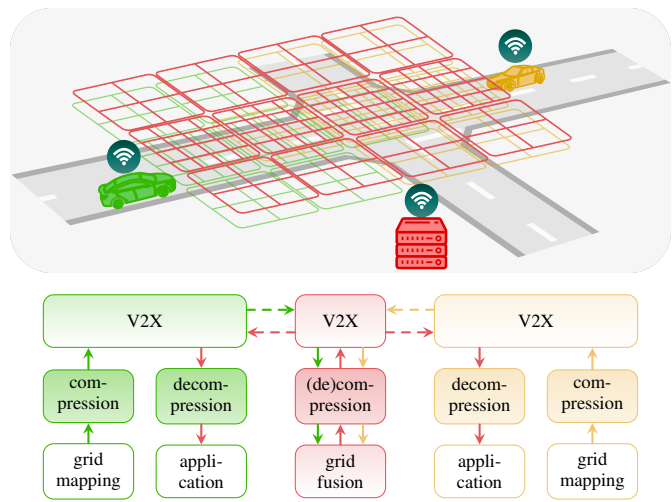}
    \caption{V2X use case example for the presented approach. The analyzed compression step, including decompression, enables V2X applications for local grid maps (green and yellow). Then, the grid fusion of multiple grid maps from multiple CAVs can be done on the server. The fused grid (red) can then be returned to the CAVs for various applications. The wireless transmission is indicated with the dashed arrows.}
    \label{fig:teaser}
\end{figure}

Grid-map-based approaches divide the environment into separate areas, called cells.
Respectively, each cell represents the state of a specific location relative to a global reference or the vehicle.
For example, a cell commonly contains information about whether the area is currently occupied or free~\cite{thrun2005probabilistic}.
Semantic or velocity information may also be included~\cite{richter2022mapping, nuss2018random}.
In contrast to object-based tracking approaches, grid maps usually make few to no model assumptions and can thus provide essential information for safety-critical tasks.
Within the context of intelligent vehicles, however, one downside of fewer assumptions is that large areas must be covered, which may only be sparsely filled with information.

The authors of~\cite{wodtko23} proposed an approach that reduces the number of cells by considering reachability and required resolutions, which has proven efficient, especially in changing situations for autonomous vehicles.
While this reduces the overall memory size required to represent a grid map, it still focuses on the application on a single processing unit. 
Thus, floating point precision and fusion stages are set to be most efficient on the compute platform in terms of computational expense and processing time.
In contrast, serialization and transmission times are not considered.

Connecting an intelligent vehicle to its surrounding infrastructure and other traffic participants, vehicle-to-everything (V2X) communication is well established and essential in the development of new autonomous driving technologies~\cite{vanKempen23}.
\ifdefined\DOUBLEBLIND
Not only does it allow for the mere exchange of information between connected autonomous vehicles (CAVs), but it also facilitates computational power requirements and in-vehicle behavior planning through novel aspects, like function offloading and cooperative planning strategies~\cite{saeik2021, hult18}.
\else
Not only does it allow for the mere exchange of information between connected autonomous vehicles (CAVs), but it also facilitates computational power requirements and in-vehicle behavior planning through novel aspects, like function offloading and cooperative planning strategies~\cite{saeik2021, dehler25, mertens22, hult18}.
\fi
However, as mentioned earlier, grid-map-based fusion is not yet available for V2X-based approaches.
To the best of our knowledge, there is a lack of an in-depth analysis discussing the possible application of compression and precision reduction methods, potentially decreasing the data size and making grid-map representations V2X-ready.

This work investigates grid-map transmission in two applications: intra-vehicle communication and V2X-based transmission. 
As an example, Fig.~\ref{fig:teaser} presents an overview of the idea for V2X-based transmission.
For grid-map transmission, we combine well-known compression algorithms with a quantization method, and examine the possible applicability in the context of intelligent vehicles.
After presenting a straightforward quantization approach tailored to grid-map representations, the essential properties of the different transmission systems are outlined, and an objective function describing the overall latency is formally derived.
Five different compression algorithms are then investigated in combination with the quantization method in a real-world test setup.
Finally, based on the results, guidelines for grid-map transmission are presented.

Summarizing the contribution of this work, we propose:
\begin{itemize}
    \item our novel grid-map quantization approach (Sec.~\ref{sec::quantization});
    \item an in-depth investigation of compression algorithms with real-world data of CAVs (Secs.~\ref {sec:analysis} and \ref{sec:evaluation}); and
    \item guidelines for grid-map transmission in the context of intelligent vehicles (Sec.~\ref{sec:conclusion}).
\end{itemize}
\section{Background}
Grid maps were initially introduced in general robotic applications as a method for representing a robot's close environment.
The area of each grid map is usually square, with a cell resolution of a few centimeters. 
However, traditional grid maps for autonomous driving have several limitations due to the larger area that must be covered.
Modeling this area with a grid map with cell sizes of centimeters creates vast amounts of data.
Consequently, grid-map processing nowadays is commonly done on Graphics Processing Units (GPUs). 

The Adaptive Patched Grid Map (APGM) offers a solution for this problem by adding another level between the environment and the cell level that is called patches~\cite{wodtko23}.
These square patches can be dynamically located in areas relevant for the driving task, where each patch is significantly smaller than traditional grid maps.
Additionally, the granularity of cells, i.e., the cell sizes, in each patch can differ depending on the respective relevance of the covered area.
This results in an efficient solution to cover the critical areas for intelligent vehicles, e.g., the streets around and especially ahead.
However, the resulting grid maps are still large in volume, especially if sent through a wireless transmission medium for a V2X application. 

To overcome the bottleneck of the huge amount of data that is generated and exchanged between multiple processing units within a CAV, but also between multiple CAVs and servers, compression techniques have already been proposed for common data types and representations in autonomous driving.
Lidar sensors create huge amounts of data, since each generated point needs to be stored. 
A thorough survey of multiple techniques for compressing point clouds is conducted in~\cite{roriz24}.

With a similar objective to the one we pursue in this paper, some approaches have also proposed compression techniques for grid map data.
In~\cite{nelson15}, a compression technique is proposed, where the grid map's resolution is modified depending on relevant information.
However, the resolution modification evidently results in a lossy compression method, with the information stored in the grid map also being reduced.
Additionally, the evaluation does not focus on the transmission of grid maps, especially not for V2X-based communication technologies with smaller bandwidths.
An optimization approach for compressing grid maps emphasizing the available communication bandwidth is proposed in~\cite{larsson2025}.
The compression uses quadtree structures to prune specific areas of grid maps.
The optimization also involves grid-map quantization, similar to what we analyze in this paper.
However, through pruning the original data is not entirely reproducible.
Thus, lossless compression is still not considered.

In the state of the art, there is a lack of lossless compression algorithms for grid-map data in the context of distributed intelligent transportation systems that we address with this work.
Additionally, we also consider the combination of lossless compression and lossy quantization.
\section{Compression Techniques}
This section summarizes the different compression methods that are applied and analyzed for an efficient transmission of grid maps.
First, the applied lossless compression algorithms, namely LZ4~\cite{collet11}, Zstandard (Zstd)~\cite{zstandard}, Run-Length Encoding (RLE)~\cite{golomb66}, a combination of RLE and LZ4 that we call RLZ4, and PNG compression are briefly presented.

Afterward, quantization is discussed as a well-known technique for decreasing the data size.
Other lossy compression algorithms are explicitly excluded in this analysis, as these algorithms would require deciding to what degree information loss is acceptable.
Since that decision is application-specific, we limit lossy compression within our pipeline to a reduction of precision via quantization.
The quantization approach is then also combined with the lossless compression algorithms.
For our compression approaches, the algorithms operate directly on serialized bitstreams, independent of the semantic structure or domain-specific meaning of the data.

\subsection{Lossless compression algorithms}
In the following, each algorithm used for lossless data compression is introduced.

\subsubsection{LZ4}
LZ4 is a compression algorithm from the Lempel-Ziv 77 (LZ77) family, a collection of compression algorithms using dictionary-based coding, founded on~\cite{ziv77, ziv78}.
In principle, it compresses by searching for patterns in the data that need to be compressed and replacing them with a smaller representation.
The algorithm is known to be very fast for compression and decompression with a small trade-off in the compression ratio~\cite{maulidina24}.
Additionally, an acceleration parameter $a_\text{LZ4}$ can be used to control this trade-off.
In the following, we use the notation LZ4$_{a_\text{LZ4}}$ to indicate the acceleration parameter.

\subsubsection{Zstandard}
Another member of the LZ77 family is the Zstd compression algorithm.
In addition to LZ77 compression, Zstd employs entropy encoding, specifically Finite State Entropy, a fast, table-driven form of Asymmetric Numeral Systems~\cite{duda13}.
Through this combination, Zstd achieves a fast compression time and high compression ratios~\cite{zstandard}.
For Zstd, a compression level parameter $c_\text{Zstd}$ can be used to regulate the trade-off between compression and speed.
In the following, we notate Zstd$_{c_\text{Zstd}}$ to indicate the compression level parameter.

\subsubsection{Run-Length Encoding}
The simplest compression method that we use in our analysis is RLE.
As the name suggests, RLE compresses data by replacing sequences of repeated symbols with the symbol and its number of direct repetitions~\cite{golomb66}.
RLE is especially efficient for data with many value repetitions.
However, e.g., in noisy data, RLE can possibly increase the data size, making its application limited.

\subsubsection{Hybrid RLE+LZ4 (RLZ4)}
We analyze a hybrid approach, in which we use RLE as a preprocessing step before employing LZ4.
The motivation behind this design is that RLE replaces long sequences of repeated data with shorter representations, which can be exploited by an algorithm optimized for compression and decompression speed, such as LZ4.
Unlike algorithms which use Huffman or Finite State Entropy, LZ4 relies on a dictionary matching stage with a greedy approach.
This preprocessing step should enhance the matching capabilities of LZ4.
The additional processing overhead is offset by the reduced message size, particularly in scenarios with bandwidth constraints or high latency.

\subsubsection{Portable Network Graphics}
With grid maps having a similar structure to grayscale images, we also investigate the application of PNG, a lossless compression algorithm, that is designed for images.
PNG compression employs two steps, i.e., filtering and compression, where in the second step, the deflate algorithm is used, combining a greedy LZ77 approach with Huffman coding~\cite{sayood2003}.

\subsection{Grid-Map Quantization}
\label{sec::quantization}

\ifdefined\DOUBLEBLIND
The grid map representation used in this work is an extended version of the APGM~\cite{wodtko23}.
The extension involves a cell representation based on Subjective Logic (SL).
\else
After the initial publication of \cite{wodtko23}, we extended the APGM framework with a cell representation based on Subjective Logic (SL).
\fi
Hence, we briefly introduce the underlying data representation before focusing on the application of quantization.
For a detailed introduction to SL, we refer to \cite{josang2016subjective}.

The grid maps considered in this work are limited to occupancy information. For each cell, the probabilistic occupancy state is represented in a so-called binomial opinion $\omega = (b, d, u, a)$. 
This opinion, a basic representation in SL, collects evidence for occupancy in the belief mass $b$. 
Similarly, evidence for the counter-event, i.e., free space, is represented in the disbelief $d$.
The uncertainty mass $u$ models the lack of evidence and is linked to the other two masses through the additivity requirement $b + d + u = 1$.
Consequently, it is sufficient to store two out of the three masses to represent the opinion. Moreover, each of the three masses is limited to the range $[0, 1]$.
The base rate $a$ reflects the prior of a cell being occupied.
\ifdefined\DOUBLEBLIND
Since this occupancy prior is the same for all cells, our adapted APGM framework optimizes memory requirements by storing it only once, rather than per cell.
Thus, the memory representation of a cell consists solely of the two masses $b$ and $d$.
\else
Since this occupancy prior is the same for all cells, the APGM framework optimizes memory requirements by storing it only once, rather than per cell.
Thus, the memory representation of a cell within the APGM framework consists solely of the two masses $b$ and $d$.
\fi
Assuming storage via 32-bit floating point representations, a single cell allocates 8 bytes. 

While a floating-point representation is beneficial for directly applying operators defined by SL, our focus here is on data transmission; thus, precision and data size are of interest.
Therefore, we apply uniform quantization to the masses stored per cell to reduce the memory requirements.
More specifically, we exploit that the masses are limited to the range $[0, 1]$.
Aiming for a $1$-byte quantized representation, this range is divided into $255$ equally spaced quantization steps.
The resulting precision is defined by the quantization step $\Delta$ given by
\begin{equation}
    \Delta = \frac{1}{255} \approx 0.00392.
\end{equation}
Although quantization does not allow recovering the original mass, the maximum error introduced through the operation is bounded and given by:
\begin{equation}
    E_{max} = \frac{\Delta}{2} \approx 0.00196.
\end{equation}
In other words, the applied quantization reduces memory requirements by \SI{75}{\percent} while introducing a maximum error of approximately \SI{0.2}{\percent} for the represented mass.
By selecting a different quantization step, the trade-off between memory reduction and resulting quantization error can be adjusted according to application-specific needs.
For most applications, however, we consider a maximum quantization error of $~0.2\%$ as acceptable.
Thus, we include the quantization as a viable technique to reduce the data size for efficient grid-map transmission.
\section{Analysis of Efficient Transmission Methods}\label{sec:analysis}
Given the introduced compression techniques, we conduct a thorough analysis of these algorithms applied to grid maps.

\ifdefined\DOUBLEBLIND
The main objective of this work is to efficiently transport the grid maps.
\else
The main objective of this work is to efficiently transport the APGMs.
\fi
Thus, we analyze the time needed for data compression and decompression, serialization and deserialization, and transport, considering typical bandwidths for intra-vehicle and V2X communication.
The underlying framework we use for sending the grid maps is \ros.
The grid maps are sent as \ros messages, with each patch coded as a bytestream in the message.

\ifdefined\DOUBLEBLIND
The data used for our analysis was recorded in our autonomous test vehicle, with the grid maps representing a typical suburban area.\footnote{\label{doubleblind}Details on the experimental platform with respective references have been removed to ensure a double-blind review. They will be added for the final version.}
\else
The data used for our analysis was recorded in our test vehicle, with the grid maps representing the area around our real-world test site in Ulm, Lehr~\cite{buchholz22}.
\fi
Close to the vehicle, a cell size of $\SI{10}{\centi\meter}$ is used.
To reduce the number of cells, the cell size increases to $\SI{20}{\centi\meter}$ and further to $\SI{40}{\centi\meter}$ with increasing distance from the vehicle.
The resulting grid maps contain, on average, approximately $\num{350000}$ cells.
The experiments were conducted on an \textit{AMD~Ryzen~9~7950X~16-Core} CPU with a maximum frequency of $\SI{5.7}{\giga\hertz}$.
It is worth mentioning that we currently do not support multi-threaded execution.
To minimize noise errors, we ran each analysis 10 times and recorded the average values.
Each run is represented by more than $\num{2000}$ grid maps.

\subsection{Full Compression vs. Patch-wise Compression}
\begin{figure}[t]
    \centering
    \vspace{3pt}
    \resizebox{\columnwidth}{!}{%
\begin{tikzpicture}[baseline=(current bounding box.center), double distance=0.75pt]

\pgfdeclarelayer{background layer}
\pgfsetlayers{background layer,main}

\def\xdist{.8cm};
\def\ydist{.2cm};



\def\xNum{2}
\def\yNum{2}
\foreach \px in {0,...,\xNum}{
    \draw[dashed, line width=1, draw=muted-gray!80] (\px * \patchSize, 0 * \patchSize - \gridMargin) -- (\px * \patchSize, \yNum*\patchSize + \gridMargin);
}
\foreach \py in {0,...,\yNum} {
    \draw[dashed, line width=1, draw=muted-gray!80] (0 * \patchSize - \gridMargin, \py * \patchSize) -- (\xNum * \patchSize + \gridMargin, \py*\patchSize);
}

\foreach \px in {0}{
    \foreach \py in {0,1}
    {
        \ifnum\py=0
          \def\patchColor{dark-red};
        \else
          \def\patchColor{light-red};
        \fi

        \drawPatch{4}{(\px * \patchSize + 0.5*\patchSize,\py * \patchSize + 0.5*\patchSize)}{draw=\patchColor};
    }
}
\foreach \px in {1}{
    \foreach \py in {0,1}
    {
        \ifnum\py=0
          \def\patchColor{med-brown};
        \else
          \def\patchColor{light-brown};
        \fi
        \drawPatch{2}{(\px * \patchSize + 0.5*\patchSize,\py * \patchSize + 0.5*\patchSize)}{draw=\patchColor};
    }
}

\node (comp1) [rounded box, top color=muted-blue!60, bottom color=muted-blue!40, minimum height=.75cm, outer sep=2pt] at (3*\patchSize, -.5\patchSize) {compress};
\node (comp2) [rounded box, top color=muted-blue!60, bottom color=muted-blue!40, minimum height=.75cm, outer sep=2pt] at (3*\patchSize, .5\patchSize) {compress};
\node (comp3) [rounded box, top color=muted-blue!60, bottom color=muted-blue!40, minimum height=.75cm, outer sep=2pt] at (3*\patchSize, 1.5\patchSize) {compress};
\node (comp4) [rounded box, top color=muted-blue!60, bottom color=muted-blue!40, minimum height=.75cm, outer sep=2pt] at (3*\patchSize, 2.5\patchSize) {compress};

\node (help1) at (.5\patchSize, 0) {};
\node (help2) at (2*\patchSize, .5*\patchSize) {};
\node (help3) at (2*\patchSize, 1.5*\patchSize) {};
\node (help4) at (0.5\patchSize, 2*\patchSize) {};

\def\arrowshorten{1pt};
\draw[thick, arrow,>=stealth,shorten <=\arrowshorten] (help1) |- (comp1);
\draw[thick, arrow,>=stealth] (help2) -- (comp2);
\draw[thick, arrow,>=stealth] (help3) -- (comp3);
\draw[thick, arrow,>=stealth,shorten <=\arrowshorten] (help4) |- (comp4);

\node (ser) [rounded box, minimum height=15*\patchSizeCm, minimum width=.8cm, align=center, top color=muted-gold!70, bottom color=muted-gold!50, outer sep=2pt] at (4.5*\patchSize, \patchSize) {\rotatebox{90}{serialize}};

\draw[thick, arrow,>=stealth] (comp1.east) -- (ser);
\draw[thick, arrow,>=stealth] (comp2.east) -- (ser);
\draw[thick, arrow,>=stealth] (comp3.east) -- (ser);
\draw[thick, arrow,>=stealth] (comp4.east) -- (ser);

\node (deser) [rounded box, minimum height=15*\patchSizeCm, minimum width=.8cm, align=center, top color=muted-green!60, bottom color=muted-green!40, outer sep=2pt] at (7.0*\patchSize, \patchSize) {\rotatebox{90}{deserialize}};

\node[align=center, font=\footnotesize, muted-darkred] at ($(ser)!0.5!(deser)$) {efficient\\transmission};
\draw[thick, arrow,>=stealth, muted-darkred] (ser) -- (deser);

\node (dec1) [rounded box, top color=muted-purple!60, bottom color=muted-purple!40, align=center, minimum height=.75cm, outer sep=2pt] at (8.5*\patchSize, -.5\patchSize) {de-\\compress};
\node (dec2) [rounded box, top color=muted-purple!60, bottom color=muted-purple!40, align=center, minimum height=.75cm, outer sep=2pt] at (8.5*\patchSize, .5\patchSize) {de-\\compress};
\node (dec3) [rounded box, top color=muted-purple!60, bottom color=muted-purple!40, align=center, minimum height=.75cm, outer sep=2pt] at (8.5*\patchSize, 1.5\patchSize) {de-\\compress};
\node (dec4) [rounded box, top color=muted-purple!60, bottom color=muted-purple!40, align=center, minimum height=.75cm, outer sep=2pt] at (8.5*\patchSize, 2.5\patchSize) {de-\\compress};

\draw[thick, arrow,>=stealth] (deser) -- (dec1.west);
\draw[thick, arrow,>=stealth] (deser) -- (dec2.west);
\draw[thick, arrow,>=stealth] (deser) -- (dec3.west);
\draw[thick, arrow,>=stealth] (deser) -- (dec4.west);

\def\xOffset{9.5*\patchSize};
\foreach \px in {0,...,\xNum}{
    \draw[dashed, line width=1, draw=muted-gray!80] (\xOffset + \px * \patchSize, 0 * \patchSize - \gridMargin) -- (\xOffset + \px * \patchSize, \yNum*\patchSize + \gridMargin);
}
\foreach \py in {0,...,\yNum} {
    \draw[dashed, line width=1, draw=muted-gray!80] (\xOffset + 0 * \patchSize - \gridMargin, \py * \patchSize) -- (\xOffset + \xNum * \patchSize + \gridMargin, \py*\patchSize);
}

\foreach \px in {0}{
    \foreach \py in {0,1}
    {
        \ifnum\py=0
          \def\patchColor{dark-red};
        \else
          \def\patchColor{light-red};
        \fi

        \drawPatch{4}{(\xOffset + \px * \patchSize + 0.5*\patchSize,\py * \patchSize + 0.5*\patchSize)}{draw=\patchColor};
    }
}
\foreach \px in {1}{
    \foreach \py in {0,1}
    {
        \ifnum\py=0
          \def\patchColor{med-brown};
        \else
          \def\patchColor{light-brown};
        \fi
        \drawPatch{2}{(\xOffset + \px * \patchSize + 0.5*\patchSize,\py * \patchSize + 0.5*\patchSize)}{draw=\patchColor};
    }
}

\node (help1-1) at (\xOffset + 1.5\patchSize, 0) {};
\node (help2-1) at (\xOffset, .5*\patchSize) {};
\node (help3-1) at (\xOffset, 1.5*\patchSize) {};
\node (help4-1) at (\xOffset + 1.5\patchSize, 2*\patchSize) {};

\draw[thick, arrow,>=stealth, shorten >=\arrowshorten,shorten <=\arrowshorten] (dec1) -| (help1-1);
\draw[thick, arrow,>=stealth, shorten <=\arrowshorten] (dec2) -- (help2-1);
\draw[thick, arrow,>=stealth, shorten <=\arrowshorten] (dec3) -- (help3-1);
\draw[thick, arrow,>=stealth, shorten >=\arrowshorten,shorten <=\arrowshorten] (dec4) -| (help4-1);

\end{tikzpicture}
}
    \caption{Simple overview of patch-wise compression. Each patch, indicated with different colors, is compressed separately before the whole \ros message is serialized. For reconstruction, the whole \ros message is deserialized, and each patch is again decompressed separately.}
    \label{fig:patch-wise}
\end{figure}
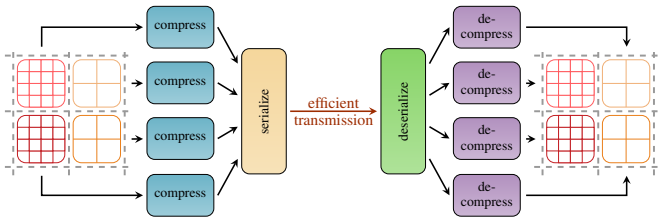

With the grid maps stored as \ros messages, the compression algorithms can be applied on the serialized data of the whole \ros message or separately on each patch of the grid map.
When compressing the whole \ros message at once, the serialization needs to be done prior to the compression, while for compressing each patch separately, first the compression is done and then serialization as a second step.
This patch-wise compression is possible because the grid-maps are already stored as bytestreams.
An illustrative overview for patch-wise compression of an APGM with four different patches is seen in Fig.~\ref{fig:patch-wise}.

\begin{figure}[t]
    \centering
    \vspace{3pt}
    \begin{tikzpicture}

  \pgfplotstableread[col sep=comma]{img/eval_full_vs_patch/data_2.csv}\datatable

  \def\xticks {0.8,1.2,1.8,2.2,2.8,3.2,3.8,4.2,4.8}
  \def\xlabels {{LZ4},{LZ4$_P$},{RLE},{RLE$_P$},{RLZ4},{RLZ4$_P$},{Zstd},{Zstd$_P$}}

  \begin{axis}[
    ybar stacked,
    bar width=15pt,
    ylabel={Computation time $[\SI{}{\milli\second}]$},
    ylabel style={font=\small},
    height=5cm,
    width=\linewidth,
    legend style={at={(0.2,0.95)}, anchor=north, legend columns=1, font=\scriptsize, inner sep=1pt},  
    xtick=\xticks,
    xticklabels={{LZ4\ \ },{\ \ LZ4$_P$},{RLE\ \ },{\ RLE$_P$},{RLZ4\ \ },{\ \ \ RLZ4$_P$},{Zstd\ \ },{\ \ Zstd$_P$},{\ \ PNG$_P$}},
    xticklabel style={font=\scriptsize},
    xtick pos=bottom,
    ymin=0,
    ymajorgrids=true,
    reverse legend,
  ]

  \addplot[muted-gold, fill=muted-gold!\coloralpha] table[x=pos, y=ser]{\datatable};
  \addplot[muted-blue, fill=muted-blue!\coloralpha] table[x=pos, y=comp]{\datatable};
  \addplot[muted-purple, fill=muted-purple!\coloralpha] table[x=pos, y=decomp]{\datatable};
  \addplot[muted-green, fill=muted-green!\coloralpha] table[x=pos, y=deser]{\datatable};

  \addplot[muted-blue, fill=muted-blue!\coloralpha] table[x=pos, y=comp_p]{\datatable};
  \addplot[muted-gold, fill=muted-gold!\coloralpha] table[x=pos, y=ser_p]{\datatable};
  \addplot[muted-green, fill=muted-green!\coloralpha] table[x=pos, y=deser_p]{\datatable};
  \addplot[muted-purple, fill=muted-purple!\coloralpha] table[x=pos, y=decomp_p]{\datatable};

  \legend{compression, serialization, deserialization, decompression}

  \end{axis}
\end{tikzpicture}
    \caption{Comparison of the summed time of (i) serialization, (ii) compression, (iii) decompression, and (iv) deserialization of the whole \ros message to the patch-wise compression pipeline with (i) patch-wise compression, (ii) serialization, (iii) deserialization, and (iv) patch-wise decompression. Patch-wise compression is indicated with the subscript $P$.}
    \label{fig:fullVsPatch}
\end{figure}
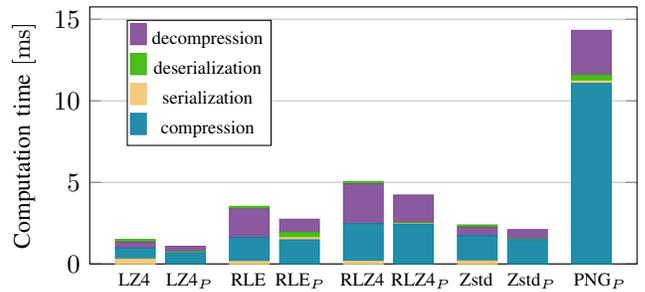

Figure~\ref{fig:fullVsPatch} shows the \ete times of the two different approaches for the different lossless compression techniques.
The time measured for the whole message is calculated by the sum of the time for serializing the \ros message, compression of the serialized message, and then decompression and deserialization.
The time for the patch-wise approach is the sum of the time for compressing each patch, serializing the \ros message with the compressed patches, then deserializing this message, and decompressing each patch, as it is also shown in Fig.~\ref{fig:patch-wise}.
The different order of the \mbox{(de-)serialization} and \mbox{(de-)compression} steps for the two approaches is also visualized in Fig.~\ref{fig:fullVsPatch}.

For the results in Fig.~\ref{fig:fullVsPatch}, we use the default parameters for the compression algorithms, i.e., LZ4$_1$ and Zstd$_3$.
The plot shows that the overall time for patch-wise compression is smaller than compressing the full \ros message.
This is because the data that needs to be serialized is comparatively smaller for patch-wise compression, as we serialize the already compressed data.
The more efficient serialization and deserialization can also be seen in Fig.~\ref{fig:fullVsPatch}, where the corresponding bars, in colors yellow and green, respectively, are significantly smaller for the patch-wise approach.
The additional overhead of using the compression algorithms multiple times for each patch is relatively small compared to the difference in the serialization time.

For PNG compression, we have only considered the patch-wise approach.
This is a consequence of normal images having a uniform pixel grid, while the cell sizes within an APGM can differ across patches.

When comparing the different algorithms, it is visible that LZ4$_P$ and Zstd$_P$ are significantly more efficient compared to RLE$_P$, RLZ4$_P$ and PNG$_P$.
This result is also consistent for the following analyses.
As a consequence of these results, we only consider the patch-wise compression and the compression algorithms LZ4$_P$ and Zstd$_P$ for the remainder of this paper.

Note that the analysis was conducted on equal hardware architectures.
For different architectures, e.g., ARM or x86, a change in endianness requires consideration regarding compression, as is the case for all V2X applications.
Therefore, this consideration falls outside the scope of this paper.

\subsection{Compression Algorithm Parameters}

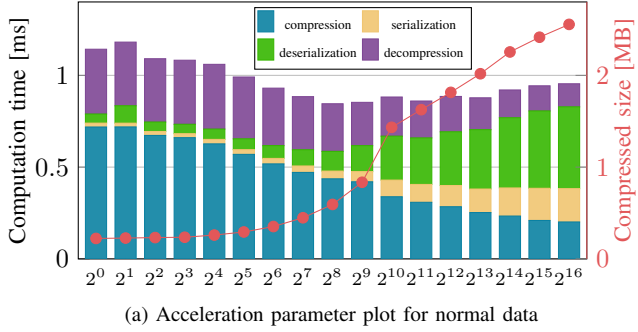
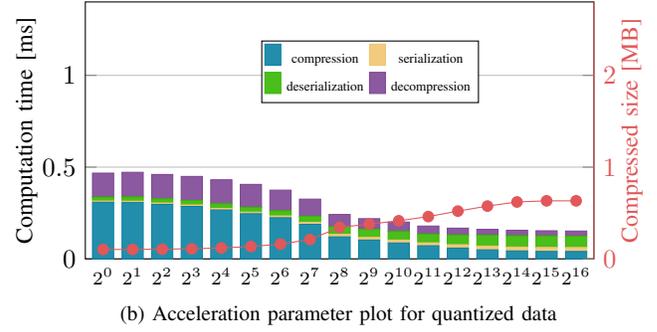
\begin{figure*}[t]
    \centering%
    \hfill%
    \subfloat[Acceleration parameter plot for normal data]{%
        \label{fig:tbd_for_subfigures}%
  \def\datafile{img/eval_lz4/data_lz4.csv}%
  \begin{tikzpicture}

    \pgfplotstableread[col sep=comma]{\datafile}\datatable

    \IfSubStr{\datafile}{quantized}{%
      \def\position{(0.56,0.85)}%
    }{%
      \def\position{(0.56,0.98)}%
    }

    \def\xticks {1,2,3,4,5,6,7,8,9,10,11,12,13,14,15,16,17}
    \def\xlabels {{LZ4},{LZ4$_P$},{RLE},{RLE$_P$},{RLZ4},{RLZ4$_P$},{Zstd},{Zstd$_P$}}
    
    \def\width {\linewidth*0.46}
    \def\height {5cm}
    
    \begin{axis}[
    ybar stacked,
    bar width=8pt,
    ylabel={Computation time [ms]},
    ylabel style={font=\small},
    ylabel shift=-5pt,
    height=\height,
    width=\width,
    legend style={at={\position}, anchor=north, legend columns=2, font=\tiny, inner sep=1pt},
    xmin=0.4, xmax=17.6,
    xtick=\xticks,
    xticklabels={{$2^0$},{$2^1$},{$2^2$},{$2^3$},{$2^4$},{$2^5$},{$2^6$},{$2^7$},{$2^8$},{$2^9$},{$2^{10}$},{$2^{11}$},{$2^{12}$},{$2^{13}$},{$2^{14}$},{$2^{15}$},{$2^{16}$}},
    xticklabel style={font=\scriptsize},
    xtick pos=bottom,
    ytick pos=left,
    ymin=0, ymax=1.4,
    ymajorgrids=true,
    ]
        \addplot[muted-blue, fill=muted-blue!\coloralpha] table[x=pos, y=comp]{\datatable};
        \addplot[muted-gold, fill=muted-gold!\coloralpha] table[x=pos, y=ser]{\datatable};
        \addplot[muted-green, fill=muted-green!\coloralpha] table[x=pos, y=deser]{\datatable};
        \addplot[muted-purple, fill=muted-purple!\coloralpha] table[x=pos, y=decomp]{\datatable};
        
        \legend{compression, serialization, deserialization, decompression}
    
    \end{axis}
    
    \begin{axis}[
      muted-red,
      axis y line*=right,
      axis x line=none,
      height=\height,
      width=\width,
      ylabel={Compressed size [MB]},
      ylabel style={font=\small, muted-red},
      ylabel shift=-5pt,
      xmin=0.4, xmax=17.6,
      ymin=0, ymax=2.8
    ]
        \addplot[sharp plot,mark=*,muted-red] table [x=pos, y=sizes, col sep=comma]{\datatable};
    \end{axis}
    
\end{tikzpicture}%

    }%
    \hfill%
    \subfloat[Acceleration parameter plot for quantized data]{%
        \label{fig:tbd_for_subfigures2}%
  \def\datafile{img/eval_lz4/data_lz4_quantized.csv}%
  \begin{tikzpicture}

    \pgfplotstableread[col sep=comma]{\datafile}\datatable

    \IfSubStr{\datafile}{quantized}{%
      \def\position{(0.56,0.85)}%
    }{%
      \def\position{(0.56,0.98)}%
    }

    \def\xticks {1,2,3,4,5,6,7,8,9,10,11,12,13,14,15,16,17}
    \def\xlabels {{LZ4},{LZ4$_P$},{RLE},{RLE$_P$},{RLZ4},{RLZ4$_P$},{Zstd},{Zstd$_P$}}
    
    \def\width {\linewidth*0.46}
    \def\height {5cm}
    
    \begin{axis}[
    ybar stacked,
    bar width=8pt,
    ylabel={Computation time [ms]},
    ylabel style={font=\small},
    ylabel shift=-5pt,
    height=\height,
    width=\width,
    legend style={at={\position}, anchor=north, legend columns=2, font=\tiny, inner sep=1pt},
    xmin=0.4, xmax=17.6,
    xtick=\xticks,
    xticklabels={{$2^0$},{$2^1$},{$2^2$},{$2^3$},{$2^4$},{$2^5$},{$2^6$},{$2^7$},{$2^8$},{$2^9$},{$2^{10}$},{$2^{11}$},{$2^{12}$},{$2^{13}$},{$2^{14}$},{$2^{15}$},{$2^{16}$}},
    xticklabel style={font=\scriptsize},
    xtick pos=bottom,
    ytick pos=left,
    ymin=0, ymax=1.4,
    ymajorgrids=true,
    ]
        \addplot[muted-blue, fill=muted-blue!\coloralpha] table[x=pos, y=comp]{\datatable};
        \addplot[muted-gold, fill=muted-gold!\coloralpha] table[x=pos, y=ser]{\datatable};
        \addplot[muted-green, fill=muted-green!\coloralpha] table[x=pos, y=deser]{\datatable};
        \addplot[muted-purple, fill=muted-purple!\coloralpha] table[x=pos, y=decomp]{\datatable};
        
        \legend{compression, serialization, deserialization, decompression}
    
    \end{axis}
    
    \begin{axis}[
      muted-red,
      axis y line*=right,
      axis x line=none,
      height=\height,
      width=\width,
      ylabel={Compressed size [MB]},
      ylabel style={font=\small, muted-red},
      ylabel shift=-5pt,
      xmin=0.4, xmax=17.6,
      ymin=0, ymax=2.8
    ]
        \addplot[sharp plot,mark=*,muted-red] table [x=pos, y=sizes, col sep=comma]{\datatable};
    \end{axis}
    
\end{tikzpicture}%

    }%
    \hfill%
    \caption{Time and compressed size comparison of different acceleration parameters for LZ4 compression for both normal (left) and quantized (right) data.}
    \label{fig:lz4param}
\end{figure*}
\begin{figure*}[t]
    \centering%
    \hfill%
    \subfloat[Compression level parameter for normal data]{%
  \def\datafile{img/eval_zstd/data_full.csv}%
  \begin{tikzpicture}
    \pgfplotsset{width=7cm,compat=1.18}
    \pgfplotstableread[col sep=comma]{\datafile}\datatable

    \def\position{(0.65,0.95)}%
    \def\xticks {1,2,3,4,5,6,7,8,9,10,11,12}
    
    \def\width {\linewidth*0.46}
    \def\height {5cm}
    
    \begin{axis}[
    ybar stacked,
    bar width=8pt,
    ylabel={Computation time [ms]},
    ylabel style={font=\small},
    ylabel shift=-5pt,
    height=\height,
    width=\width,
    legend style={at={\position}, anchor=north, legend columns=2, font=\tiny, inner sep=1pt},
    xmin=0.4, xmax=12.6,
    xtick=\xticks,
    xticklabels={{-100},{-90},{-80},{-70},{-60},{-50},{-40},{-30},{-20},{-10},{0},{10}},
    xticklabel style={font=\scriptsize},
    xtick pos=bottom,
    ymin=0, ymax=12,
    ymajorgrids=true,
    ]
    
        \addplot[muted-blue, fill=muted-blue!\coloralpha] table[x=pos, y=comp]{\datatable};
        \addplot[muted-gold, fill=muted-gold!\coloralpha] table[x=pos, y=ser]{\datatable};
        \addplot[muted-green, fill=muted-green!\coloralpha] table[x=pos, y=deser]{\datatable};
        \addplot[muted-purple, fill=muted-purple!\coloralpha] table[x=pos, y=decomp]{\datatable};
        
        \legend{compression, serialization, deserialization, decompression, number 5}
    
    \end{axis}
    
    \begin{axis}[
      muted-red,
      axis y line*=right,
      axis x line=none,
      height=\height,
      width=\width,
      ylabel={Compressed size $[\SI{}{\kilo\byte}]$},
      ylabel style={font=\small},
      ylabel shift=-5pt,
      ymin=0, ymax=0.70 * 12 / 10,
      ytick={0,0.35,0.70},
      yticklabels={0,35,70},
      xmin=0.4, xmax=12.6,
    ]
        \addplot[sharp plot,mark=*,muted-red] table [x=pos, y=sizes, col sep=comma]{\datatable};
    \end{axis}
    
\end{tikzpicture}%

    }%
    \hfill%
    \subfloat[Compression level parameter for quantized data.]{%
  \def\datafile{img/eval_zstd/data_full_zstd_quantized.csv}%
  \begin{tikzpicture}
    \pgfplotsset{width=7cm,compat=1.18}
    \pgfplotstableread[col sep=comma]{\datafile}\datatable

    \def\position{(0.65,0.95)}%
    \def\xticks {1,2,3,4,5,6,7,8,9,10,11,12}
    
    \def\width {\linewidth*0.46}
    \def\height {5cm}
    
    \begin{axis}[
    ybar stacked,
    bar width=8pt,
    ylabel={Computation time [ms]},
    ylabel style={font=\small},
    ylabel shift=-5pt,
    height=\height,
    width=\width,
    legend style={at={\position}, anchor=north, legend columns=2, font=\tiny, inner sep=1pt},
    xmin=0.4, xmax=12.6,
    xtick=\xticks,
    xticklabels={{-100},{-90},{-80},{-70},{-60},{-50},{-40},{-30},{-20},{-10},{0},{10}},
    xticklabel style={font=\scriptsize},
    xtick pos=bottom,
    ymin=0, ymax=12,
    ymajorgrids=true,
    ]
    
        \addplot[muted-blue, fill=muted-blue!\coloralpha] table[x=pos, y=comp]{\datatable};
        \addplot[muted-gold, fill=muted-gold!\coloralpha] table[x=pos, y=ser]{\datatable};
        \addplot[muted-green, fill=muted-green!\coloralpha] table[x=pos, y=deser]{\datatable};
        \addplot[muted-purple, fill=muted-purple!\coloralpha] table[x=pos, y=decomp]{\datatable};
        
        \legend{compression, serialization, deserialization, decompression, number 5}
    
    \end{axis}
    
    \begin{axis}[
      muted-red,
      axis y line*=right,
      axis x line=none,
      height=\height,
      width=\width,
      ylabel={Compressed size $[\SI{}{\kilo\byte}]$},
      ylabel style={font=\small},
      ylabel shift=-5pt,
      ymin=0, ymax=0.70 * 12 / 10,
      ytick={0,0.35,0.70},
      yticklabels={0,35,70},
      xmin=0.4, xmax=12.6,
    ]
        \addplot[sharp plot,mark=*,muted-red] table [x=pos, y=sizes, col sep=comma]{\datatable};
    \end{axis}
    
\end{tikzpicture}%

    }%
    \hfill%
    \caption{Time and compressed size comparison of different compression level parameters for Zstd compression for normal (left) and quantized (right) data.}
    \label{fig:zstdparam}
\end{figure*}
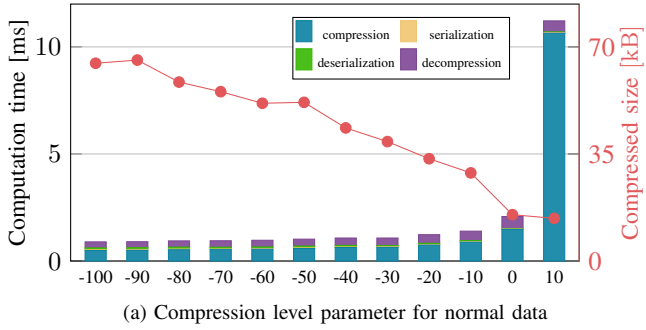
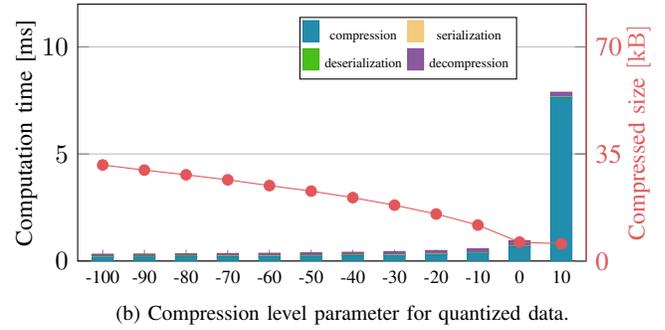

For an efficient transmission of grid maps, the significant value is the \ete time $t_\text{e2e}$ of the transport process.
The \ete time is calculated by
\begin{align}\label{eq:e2e}
    t_\text{e2e} = t_\text{comp} + t_\text{ser} + t_\text{trans} + t_\text{deser} + t_\text{decomp},
\end{align}
with the time taken for compression $t_\text{comp}$, the time for serialization $t_\text{ser}$, the time for transmitting the serialized message through the respective communication medium $t_\text{trans}$, the time for deserialization $t_\text{deser}$, and the time for decompressing the compressed data $t_\text{decomp}$.
Note that we do not include the time for quantization, even if that step is applied, since the quantization is done very efficiently directly on the GPU, and the computation time is thus negligible.

For our analysis, we do not directly measure the transmission time $t_\text{trans}$.
Instead, we calculate it using the theoretical foundation that the transmission time is the ratio of the data size $S$ to the available bandwidth $B$.
Note that additional times resulting from the communication stack and different network layers are not considered in the theoretical transmission time.
The objective of minimizing the \ete time can then be formulated as
\begin{align}
    \min t_\text{e2e} \text{,\ with\ } t_\text{trans} = \frac{S}{B} \text{,\ $B$ fixed}.
\end{align}

For the two algorithms LZ4 and Zstd, we analyze which value for the acceleration parameter and the compression level, respectively, is the best depending on the times, the size of the (compressed) data $S$, and the available fixed bandwidth $B$.

Figure~\ref{fig:lz4param} shows the individual processing times and compressed data sizes for different LZ4 acceleration parameters $a_\text{LZ4}$.
It can be observed that, up to a specific point, the computation time constantly decreases for higher $a_\text{LZ4}$, with less compression efficiency with respect to the compressed data size.
For the normal data, however, the time even increases again for $a_\text{LZ4}>2^9$.
This is because the data is compressed significantly less, which increases the serialization and deserialization time due to larger data amounts being processed in these steps.
This overhead is also visible from the color-coded contributions to the computation time in Fig.~\ref{fig:lz4param}.
For the quantized data, the increasing serialization and deserialization times are also visible, yet on a lower scale.
Thus, for quantized data, the overall times are still decreasing, resulting from relatively faster compression and decompression due to the smaller data amounts.

Figure~\ref{fig:zstdparam} shows the respective times and compressed data sizes for different Zstd compression level parameters $c_\text{Zstd}$.
For a better visualization, the plot only comprises every tenth value in the range $\left[-100, 10\right]$.
The plot shows that across the entire range, the compressed size decreases significantly for higher compression levels.
For $c_\text{Zstd}>0$, however, the compressed size decreases only slightly, while the compression time and, consequently, the overall computation time increase significantly.
Compared to the compression time, the time for serialization, deserialization, and decompression is almost negligible, which is seen in the non-visible serialization and decompression bars and the very small deserialization bars in Fig.~\ref{fig:zstdparam}.
The main advantage of Zstd compression is its higher compression ratio, resulting in significantly smaller compressed data sizes compared to LZ4 compression.

\begin{table}
\centering
\caption{Optimal parameter choice with \ete times for each considered bandwidth: $a_\text{LZ4}/c_\text{Zstd}; t_\text{e2e} [ms]$}  
\begin{tabular}{l | c | c | c | c}
    \hline
    Bandwidth & 10~Mbps & 100~Mbps & 1~Gbps & 10~Gbps \\
    \hline
    LZ4 & $1; 179.2$ & $1; 18.9$ & $1; 2.84$ & $64; 1.13$ \\
    Zstd & $8; 120.1$ & $2; 13.8$ & $-1; 2.93$ & $-62; 1.36$ \\ 
    \hline
    LZ4$^q$ & $1; 82.3$ & $1; 8.64$ & $1; 1.27$ & $128; 0.47$ \\
    Zstd$^q$ & $7; 48.8$ & $1; 5.81$ & $-1; 1.34$ & $-60; 0.56$ \\ 
    \hline
\end{tabular}
\label{table:e2etimes}
\end{table}

Given the results in Figs.~\ref{fig:lz4param} and~\ref{fig:zstdparam}, we can calculate the \ete times for each parameter.
To determine the best values, we specify common bandwidths of the considered use cases.

\ifdefined\DOUBLEBLIND
For intra-vehicle communication, we consider bandwidths of 1~Gbps and 10~Gbps, as we have in the U-Shift test vehicle~\cite{ushift2022}.
\else
For intra-vehicle communication, we consider bandwidths of 1~Gbps and 10~Gbps.
\fi
For V2X communication, we consider bandwidths for typical communication technologies, i.e., ITS-G5~\cite{etsi_es_202_663_intelligent_2009}, LTE-V2X~\cite{seo16}, and 5G-V2X~\cite{garcia21}.
For ITS-G5 and LTE-V2X, bandwidths range between 1~Mbps up to 100~Mbps~\cite{tahir21, gonzalez22}.
Considering that multiple vehicles may share the same network, we chose a suitable bandwidth of 10~Mbps for the analysis.
The bandwidth for 5G-V2X is up to 1~Gbps~\cite{szalay20}.
Again, we take into account the presence of multiple vehicles simultaneously.
Thus, we also conduct experiments with a bandwidth of 100~Mbps.

Given these bandwidths, the \ete times can be calculated with Eq.~\eqref{eq:e2e}.
Note that, even though we did not plot every value for Zstd, we still evaluated the whole range $c_\text{Zstd}\in\left[-100, 10\right]$.
Table~\ref{table:e2etimes} summarizes the optimal parameter selection for each bandwidth, i.e., where the \ete time is minimal, for LZ4 and Zstd for the normal compression and with prior quantization, which is indicated with superscript q.

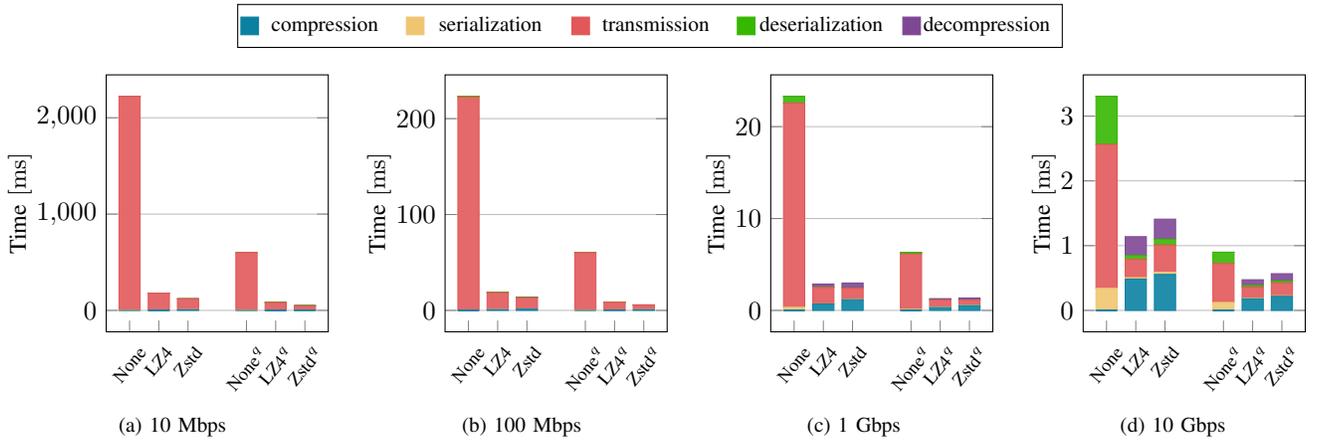
\begin{figure*}[!t]
    \centering
    \vspace{3pt}
    \begin{tikzpicture}
        \def\legendxshift{.2cm};
        \node (comp-c) [draw=muted-blue, fill=muted-blue] {};
        \node (comp) [right of=comp-c, font=\footnotesize] {compression};
        \node (ser-c) [right of=comp, xshift=\legendxshift, draw=muted-gold, fill=muted-gold] {};
        \node (ser) [right of = ser-c,  font=\footnotesize] {serialization\vphantom{p}};
        \node (trans-c) [right of=ser, xshift=\legendxshift, draw=muted-red, fill=muted-red] {};
        \node (trans) [right of = trans-c,  font=\footnotesize] {transmission\vphantom{p}};
        \node (deser-c) [right of=trans, xshift=\legendxshift, draw=muted-green, fill=muted-green] {};
        \node (deser) [right of = deser-c,  font=\footnotesize] {deserialization\vphantom{p}};
        \node (dec-c) [right of=deser, xshift=\legendxshift, draw=muted-purple, fill=muted-purple] {};
        \node (dec) [right of = dec-c,  font=\footnotesize] {decompression};

        \node[draw=black,fit=(comp-c)(dec), inner sep=1pt] (cav-group) {};
    \end{tikzpicture}
    \\
    \centering
    \hfill%
    \subfloat[10 Mbps]{%
        \label{fig:10mbps}
  \def\datafile{img/eval_algorithm_2/data_10.csv}%
  \begin{tikzpicture}

  \pgfplotstableread[col sep=comma]{\datafile}\datatable
  \begin{axis}[
    ybar stacked,
    bar width=8pt,
    ylabel={Time $[\SI{}{\milli\second}]$},
    ylabel style={font=\small},
    ylabel shift=-5pt,
    height=5cm,
    width=\linewidth*0.25,
    legend style={at={(1.5,0.98)}, anchor=north, legend columns=1, font=\scriptsize, column sep=1pt, inner sep=1pt},  
    xtick={0, 0.25, 0.5, 1, 1.25, 1.5},
    xticklabels from table={\datatable}{group},
    xticklabel style={font=\scriptsize,rotate=50},
    xtick pos=bottom,
    ymajorgrids=true,
    xmin=-.2, xmax=1.7,
  ]

  \addplot[muted-blue, fill=muted-blue!\coloralpha] table[x=pos, y=comp]{\datatable};
  \addplot[muted-gold, fill=muted-gold!\coloralpha] table[x=pos, y=ser]{\datatable};
  \addplot[muted-red, fill=muted-red!\coloralpha] table[x=pos, y=trans]{\datatable};
  \addplot[muted-green, fill=muted-green!\coloralpha] table[x=pos, y=deser]{\datatable};
  \addplot[muted-purple, fill=muted-purple!\coloralpha] table[x=pos, y=decomp]{\datatable};
  \end{axis}

\end{tikzpicture}%

    }%
    \hspace{-.2cm}
    \subfloat[100 Mbps]{%
        \label{fig:100mbps}
  \def\datafile{img/eval_algorithm_2/data_100.csv}%
  \begin{tikzpicture}

  \pgfplotstableread[col sep=comma]{\datafile}\datatable
  \begin{axis}[
    ybar stacked,
    bar width=8pt,
    ylabel={Time $[\SI{}{\milli\second}]$},
    ylabel style={font=\small},
    ylabel shift=-5pt,
    height=5cm,
    width=\linewidth*0.25,
    legend style={at={(1.5,0.98)}, anchor=north, legend columns=1, font=\scriptsize, column sep=1pt, inner sep=1pt},  
    xtick={0, 0.25, 0.5, 1, 1.25, 1.5},
    xticklabels from table={\datatable}{group},
    xticklabel style={font=\scriptsize,rotate=50},
    xtick pos=bottom,
    ymajorgrids=true,
    xmin=-.2, xmax=1.7,
  ]

  \addplot[muted-blue, fill=muted-blue!\coloralpha] table[x=pos, y=comp]{\datatable};
  \addplot[muted-gold, fill=muted-gold!\coloralpha] table[x=pos, y=ser]{\datatable};
  \addplot[muted-red, fill=muted-red!\coloralpha] table[x=pos, y=trans]{\datatable};
  \addplot[muted-green, fill=muted-green!\coloralpha] table[x=pos, y=deser]{\datatable};
  \addplot[muted-purple, fill=muted-purple!\coloralpha] table[x=pos, y=decomp]{\datatable};
  \end{axis}

\end{tikzpicture}%

    }%
    \centering%
    \hspace{-.2cm}
    \subfloat[1 Gbps]{%
        \label{fig:1gbps}
  \def\datafile{img/eval_algorithm_2/data_1000.csv}%
  \begin{tikzpicture}

  \pgfplotstableread[col sep=comma]{\datafile}\datatable
  \begin{axis}[
    ybar stacked,
    bar width=8pt,
    ylabel={Time $[\SI{}{\milli\second}]$},
    ylabel style={font=\small},
    ylabel shift=-5pt,
    height=5cm,
    width=\linewidth*0.25,
    legend style={at={(1.5,0.98)}, anchor=north, legend columns=1, font=\scriptsize, column sep=1pt, inner sep=1pt},  
    xtick={0, 0.25, 0.5, 1, 1.25, 1.5},
    xticklabels from table={\datatable}{group},
    xticklabel style={font=\scriptsize,rotate=50},
    xtick pos=bottom,
    ymajorgrids=true,
    xmin=-.2, xmax=1.7,
  ]

  \addplot[muted-blue, fill=muted-blue!\coloralpha] table[x=pos, y=comp]{\datatable};
  \addplot[muted-gold, fill=muted-gold!\coloralpha] table[x=pos, y=ser]{\datatable};
  \addplot[muted-red, fill=muted-red!\coloralpha] table[x=pos, y=trans]{\datatable};
  \addplot[muted-green, fill=muted-green!\coloralpha] table[x=pos, y=deser]{\datatable};
  \addplot[muted-purple, fill=muted-purple!\coloralpha] table[x=pos, y=decomp]{\datatable};
  \end{axis}

\end{tikzpicture}%

    }%
    \hspace{-.2cm}
    \subfloat[10 Gbps]{%
        \label{fig:10gbps}
  \def\datafile{img/eval_algorithm_2/data_10000.csv}%
  \begin{tikzpicture}

  \pgfplotstableread[col sep=comma]{\datafile}\datatable
  \begin{axis}[
    ybar stacked,
    bar width=8pt,
    ylabel={Time $[\SI{}{\milli\second}]$},
    ylabel style={font=\small},
    ylabel shift=-5pt,
    height=5cm,
    width=\linewidth*0.25,
    legend style={at={(1.5,0.98)}, anchor=north, legend columns=1, font=\scriptsize, column sep=1pt, inner sep=1pt},  
    xtick={0, 0.25, 0.5, 1, 1.25, 1.5},
    xticklabels from table={\datatable}{group},
    xticklabel style={font=\scriptsize,rotate=50},
    xtick pos=bottom,
    ymajorgrids=true,
    xmin=-.2, xmax=1.7,
  ]

  \addplot[muted-blue, fill=muted-blue!\coloralpha] table[x=pos, y=comp]{\datatable};
  \addplot[muted-gold, fill=muted-gold!\coloralpha] table[x=pos, y=ser]{\datatable};
  \addplot[muted-red, fill=muted-red!\coloralpha] table[x=pos, y=trans]{\datatable};
  \addplot[muted-green, fill=muted-green!\coloralpha] table[x=pos, y=deser]{\datatable};
  \addplot[muted-purple, fill=muted-purple!\coloralpha] table[x=pos, y=decomp]{\datatable};
  \end{axis}

\end{tikzpicture}%

    }%
    \hfill%
    
    \caption{Comparison of different \ete times including its components given different bandwidths for both normal and quantized data. The chosen optimal parameter for the plots are equivalent with Table~\ref{table:e2etimes}. The legend on the top applies to all 4 plots (a)-(d).}
    \label{fig:algorithmcomparison}
\end{figure*}
\subsection{Algorithm Comparison}
As a final analysis, we compare the different compression methods with each other.
In contrast to Fig.~\ref{fig:fullVsPatch}, the comparison is now done on the overall \ete times, including the theoretical transmission time, i.e., data size in relation to bandwidth (see Eq.~\eqref{eq:e2e}).
By including the transmission time, the size reduction is considered implicitly.

The baseline for this comparison is to use no compression technique.
Fig.~\ref{fig:algorithmcomparison} shows the \ete times for each analyzed algorithm, both normal and quantized, with the respective optimal parameter for the considered bandwidths (see Table~\ref{table:e2etimes}).
Overall, it can be seen that in each plot, both LZ4 and Zstd compression result in a lower \ete time compared to the baseline.
All plots show that the relative performance differences between the methods are similar for both variants, i.e., with and without quantization, yet on a different absolute scale.
Evidently, the \ete time drops for all methods if quantization is applied.
It is worth mentioning that even though the size reduction through quantization from 8 bytes to 2 bytes per cell is 75$\%$, the reduction of the \ete time for the compression with quantized data is lower, except for the transmission time with no compression, where the ratio is exactly $1/4$.
The lower relative reduction among the compression techniques, however, results from different compression ratios within the algorithms.
For the normal data, there are more bytes available for compression.
Thus, the compression techniques are relatively more efficient for normal data.
This result indicates that the compression ratio is lower for quantized data than for normal data.
However, the overall data size is lower for quantized data.

For the different bandwidths, it can be observed that the smaller the bandwidth, the more important the transmission time and, consequently, the data size and compression ratio.
For very small bandwidths, see Figs.~\ref{fig:10mbps} and~\ref{fig:100mbps}, the transmission time is so high that the individual processing times can be neglected and are essentially invisible in comparison to the transmission time in the plots.
For higher bandwidths, see Figs.~\ref{fig:1gbps} and~\ref{fig:10gbps}, the transmission time remains an essential factor.
However, especially the time taken for compression and decompression becomes more critical.

When comparing the algorithms with each other, the following conclusions can be drawn.
For lower bandwidths, i.e., 10~Mbps and 100~Mbps, Zstd is the best option for both normal and quantized data.
For bandwidths $\geq$1~Gbps, LZ4 offers a better solution concerning the \ete time.

We have also analyzed the exact point at which the optimal solution switches from Zstd to LZ4.
For normal data, the switch occurs at 823~Mbps, while for quantized data, it is at 737~Mbps.
It can thus be concluded that for lower bandwidths, Zstd can always be chosen, while for higher bandwidths, LZ4 is preferable.
When a bandwidth estimation is provided, the algorithm can be adjusted at  $\sim$800~Mbps.
\section{Evaluation}\label{sec:evaluation}

\ifdefined\DOUBLEBLIND
To support the theoretical analysis, we have integrated the quantization and compression algorithms into our grid-mapping framework\footnoteref{doubleblind}.
\else
To support the theoretical analysis, we have integrated the quantization and compression algorithms into our grid-mapping framework.
\fi
The setup is equivalent to the one used for data recording in Sec.~\ref{sec:analysis}, i.e., we use ROS 2 as the underlying framework.
We evaluate this setup in two scenarios that differ especially in bandwidth: V2X sharing of grid maps between a CAV and a MEC server, and intra-vehicle communication between distributed processing units (DPUs).

\subsection{V2X Application}

\ifdefined\DOUBLEBLIND
The first use-case is analyzed in our real-world test site, where we have a CAV sending data through a mobile network to a Multi-Access Edge Computing (MEC) server.
For the evaluation, we conducted multiple test drives with our experimental CAV.\footnoteref{doubleblind}

\else
The first use-case is analyzed in our real-world test site in Ulm, Lehr~\cite{buchholz22}.
More specifically, we run the tests on a Mercedes-Benz E-Class as the CAV sending data through a public 5G network to a Multi-Access Edge Computing (MEC) server.
\fi
To select suitable evaluation parameters, we determined the network characteristics of the mobile network.
The mean and standard deviation (std) of the bandwidth of our mobile network are
\begin{align}\label{eq:eval}
\begin{aligned}
    \text{mean}_\text{mobile} &= 6.36~\text{Mbps}, \\
    \text{std}_\text{mobile} &= 3.09~\text{Mbps},
\end{aligned}
\end{align}
respectively.
The high std already reflects the highly variable network conditions for V2X application. 
Given the values in Eq.~\eqref{eq:eval}, we chose the compression method parameters that we determined for 10~Mbps. 

With the results of the analysis, we already set the transmission frequency of the grid maps during the following test drives to $\SI{0.1}{\hertz}$, since we expect a high transmission time for some methods.
Thus, we do not overload the network, which would result in distorted measurements especially for non-compressed data.
For the evaluation, we recorded the mean, std, and the minimum (min) and maximum (max) values of the transmission times for the different compression versions and our baseline of no compression.
In the V2X evaluation, we transmitted approximately 200 grid maps.

The results of the V2X evaluation are summarized in the first line of Table~\ref{table:eval}.
The results are consistent with the results of our analysis.
The high variability of the network conditions is also seen in the std, min, and max values, respectively.
Comparing the different algorithms, a huge range is visible, from very high mean values of $\SI{3623.7}{\milli\second}$ for no compression to the best result for Zstd$^q$, with a mean transmission time of $\SI{201.5}{\milli\second}$.
Given the mean transmission time of approximately $\SI{200}{\milli\second}$, an average frequency of $\SI{5}{\hertz}$ may be chosen for grid map V2X application.
The V2X evaluation shows a significant improvement when comparing the baseline to our proposed methods.
These results underline the effectiveness of our approach, reducing the \ete time by a factor of
\begin{align}
\frac{\SI{3623.7}{\milli\second}}{\SI{201.5}{\milli\second}} \approx 18.
\end{align}

\subsection{Distributed Processing Units}

\ifdefined\DOUBLEBLIND
The second use case is analyzed in an experimental automated vehicle, where one distributed processing unit creates grid maps and a central computer receives the grid maps with a 10~Gbps interface connecting the two units\footnoteref{doubleblind}.
\else
The second use case is analyzed in a test setup similar to the U-Shift test vehicle~\cite{ushift2022}, where one distributed processing unit creates grid maps and a central computer receives the grid maps~\cite{ushift2025}, with a 10~Gbps interface connecting the two units.
\fi
Our evaluation is conducted in a manner equivalent to the V2X evaluation.
However, we now use the same rosbag file that we used in the analysis instead of creating the grid maps live.
For this analysis, we evidently chose the parameters that we have observed in the theoretical analysis with 10~Gbps.

The results are summarized in the second line of Table~\ref{table:eval}.
It can be observed that, due to additional overheads resulting from the communication pipeline, the results differ from the theoretical analysis.
For normal data, however, both compression methods outperform the baseline of no compression, with LZ4 being the best method, confirming the results from our previous analysis.
When considering the ratio of the mean \ete time saved, the benefit of using LZ4 compression over no compression is significant:
\begin{align}
    \frac{\SI{6.43}{\milli\second}}{\SI{5.16}{\milli\second}} \approx 1.25
\end{align}
The std values show that the 10~Gbps interface is more stable than the V2X network, with some outliers seen in the min and max values.
After observing the data it can be said that these outliers also result from different sizes of the transmitted grid maps.

While quantization is useful for V2X scenarios with low bandwidths, its benefits for intra-vehicle communication are more limited.
On non-compressed data, quantization is useful, reducing the \ete time by 67\% compared to the full-resolution representation. 
However, when compression is applied, the differences become much smaller.
For LZ4, the mean times stays exactly the same at $\SI{2.11}{\milli\second}$, while Zstd results in a minor increase to $\SI{2.50}{\milli\second}$.
In comparison, full-resolution data with LZ4 compression already achieves $\SI{5.16}{\milli\second}$, while preserving complete precision.
Thus, while quantization provides a large reduction in end-to-end time, this reduction comes at the cost of precision.
For high-throughput intra-vehicle communication, the full-resolution representation with compression is preferred.

\begin{table}
\centering
\setlength{\tabcolsep}{4pt} 
\caption{Real-world evaluation for each compression method with different time values in $\left[\SI{}{\milli\second}\right]$}
\begin{tabular}{l c | c | c | c | c | c | c}
    \hline
    \multicolumn{2}{l|}{Use-Case} & None & LZ4 & Zstd & None$^q$ & LZ4$^q$ & Zstd$^q$ \\
    \hline
    \multirow{4}{*}{V2X}& mean & $3623.7$ & $567.8$ & $304.8$ & $1450.2$ & $239.6$ & $201.5$ \\
    & std & $3353.1$ & $470.7$ & $152.8$ & $1388.0$ & $85.5$ & $129.03$ \\
    & min & $1055.3$ & $212.6$ & $123.3$ & $301.7$ & $147.1$ & $66.8$ \\
    & max & $15159$ & $2160.6$ & $830.5$ & $7394.5$ & $493.8$ & $629.9$ \\
    \hline
    \multirow{4}{*}{DPU}& mean & $6.43$ & $5.16$ & $6.29$ & $2.11$ & $2.11$ & $2.50$ \\ 
    & std & $2.55$ & $1.14$ & $1.36$ & $0.36$ & $0.44$ & $0.74$ \\
    & min & $1.3$ & $1.34$ & $1.37$ & $0.44$ & $0.59$ & $0.76$ \\
    & max & $14.79$ & $9.47$ & $15.48$ & $3.75$ & $8.41$ & $11.54$ \\
    \hline
\end{tabular}
\label{table:eval}
\end{table}
\section{Conclusion and Future Work}\label{sec:conclusion}
In this paper, we have presented and analyzed the application of various compression algorithms to grid-map data.
The theoretical analysis and real-world evaluation demonstrated the effectiveness of data compression for the use cases of intra-vehicle and V2X-based communication.
With this, we summarize the following guidelines for efficient grid-map transmission:
\begin{itemize}
    \item If grid maps are present as patched grid maps, patch-wise compression outperforms the compression of the whole grid map.
    \item Quantization is the best solution to significantly decrease the transmission time, if a trade-off between speed and resolution is acceptable.
    \item Overall, LZ4 compression is the best alternative for a broad spectrum of bandwidths.
    \item If a bandwidth estimation is possible, Zstd is the better option for lower bandwidth applications.
    \item Real-time V2X-based applications with lower bandwidths are only possible if the combination of compression and quantization is used.
\end{itemize}

In our future work, we want to improve further the compression techniques, including frame-by-frame approaches similar to video codecs.
Lastly, we will continue to evaluate how intelligent vehicles benefit from the new set of applications enabled through efficient grid-map transmission.

\balance
\bibliographystyle{utility/IEEEtran-et-al}
\bibliography{biblio}

\end{document}